\documentclass[letterpaper]{article} 
\usepackage{aaai24}  
\usepackage{times}  
\usepackage{helvet}  
\usepackage{courier}  
\usepackage[hyphens]{url}  
\usepackage{graphicx} 
\usepackage{natbib}  
\usepackage{caption} 
\usepackage{algorithm}
\usepackage{algorithm}
\usepackage{algpseudocode}
\usepackage{amsmath}
\usepackage{amssymb}
\usepackage{booktabs}
\usepackage[table,xcdraw]{xcolor}
\usepackage{caption}
\usepackage{subcaption}
\usepackage{multirow}
\usepackage{cleveref}
\usepackage[most]{tcolorbox}

\usepackage{newfloat}
\usepackage{listings}
\crefname{figure}{Fig.}{Figs.}

\DeclareCaptionStyle{ruled}{labelfont=normalfont,labelsep=colon,strut=off} 
\floatstyle{ruled}
\newfloat{listing}{tb}{lst}{}
\floatname{listing}{Listing}
\title{
Context-Aware Prompt Tuning for Vision-Language Model with Dual-Alignment}
\author{
    Hongyu Hu \textsuperscript{\rm 1,\rm 2 *},
    Tiancheng Lin \textsuperscript{\rm 1 *},
    Jie Wang \textsuperscript{\rm 2},
    Zhenbang Sun \textsuperscript{\rm 2},
    Yi Xu \textsuperscript{\rm 1 \dag}
}

\affiliations {
    \textsuperscript{\rm 1}Shanghai Jiao Tong University\\
    \textsuperscript{\rm 2}ByteDance Inc\\
    mathewcrespo@sjtu.edu.cn
}

\usepackage{bibentry}

\begin{document}

\maketitle

\begin{abstract}
Large-scale vision-language models (VLMs), $e.g.$, CLIP, learn broad visual concepts from tedious training data, showing superb generalization ability.
Amount of prompt learning methods have been proposed to efficiently adapt the VLMs to downstream tasks with only a few training samples. 
We introduce a novel method to improve the prompt learning of vision-language models by incorporating pre-trained large language models (LLMs), called \textbf{Du}al-\textbf{Al}igned \textbf{P}rompt \textbf{T}uning (\textsc{DuAl-PT}). 
Learnable prompts, like CoOp, implicitly model the context through end-to-end training, which are difficult to control and interpret.
While explicit context descriptions generated by LLMs, like GPT-3, can be directly used for zero-shot classification, such prompts are overly relying on LLMs and still underexplored in few-shot domains. 
With \textsc{DuAl-PT}, we propose to learn more context-aware prompts, benefiting from both explicit and implicit context modeling. 
To achieve this, we introduce a pre-trained LLM to generate context descriptions, and we encourage the prompts to learn from the LLM's knowledge by alignment, as well as the alignment between prompts and local image features. 
Empirically, \textsc{DuAl-PT} achieves superior performance on 11 downstream datasets on few-shot recognition and base-to-new generalization.
Hopefully,  \textsc{DuAl-PT} can serve as a strong baseline. Code will be available.

\end{abstract}

\section{Introduction}
Pre-trained vision-language models (VLMs), like CLIP~\cite{clip} and ALIGN~\cite{align}, have achieved remarkable success in learning broad visual concepts. 
These models are trained on web-scale image-text pairs to learn aligned representations of image and text through contrastive loss. 
By providing specific prompts, $e.g.,$ \texttt{A photo of a \{label\}}, they can be readily applied to downstream tasks in a manner similar to pre-training --- calculating the similarities between the task-related descriptions and images encoded by the text and image encoders respectively~\cite{clip}.

Despite significant improvements in prompt engineering,  classification only based on category names is intuitively insufficient.
It overlooks diverse characteristics of categories, resulting in a `degradation' to traditional supervised learning with discrete labels.
Moreover, manual design still requires domain expertise, suffers from high variance and inevitably introduces artificial bias, making it a non-trivial task. 
As a solution, data-driven approaches, \textit{a.k.a}, the learnable prompts, are introduced to leverage the rich context of additional information for classification.
Typical works propose one single prompt~\cite{coop}, Meta-Net~\cite{cocoop} and prompt sets~\cite{proda,plot} to implicitly model the class-wise context, instance-specific context and context variance, respectively. 
While these methods have achieved improved performance, the learned context is not always useful only by a standard classification loss ($e.g.,$ the cross-entropy).
In some cases, the learned context might coincide with each other and focus on class-agnostic background~\cite{plot}.
More recently, some works~\cite{VLMGPT3,VLMGPT3_2,VLMGPT3_3} turn to the large language models (LLMs), $e.g.,$  GPT-3, to explicitly construct context descriptions as prompts, benefiting from the inherent explainability.
They hypothesize LLMs possess remarkable world knowledge on a variety of topics.
However, a concern with these works is that they may be overly reliant on the pretrained LLMs, which are not always readily available in deployment/inference, particularly for new classes.
Moreover, they primarily focus on zero-shot tasks and do not explore the potential of LLMs in few-shot domains.
Therefore, we believe a more  efficient and effective VLM adaptation should be proposed by making use of both the flexibility of learnable prompts (implicit context modeling) and abundant knowledge of LLMs (explicit context descriptions).

\begin{figure*}
    \centering 
    
    \begin{subfigure}[t]{0.2\linewidth}
	\centering
	\includegraphics[width=\linewidth]{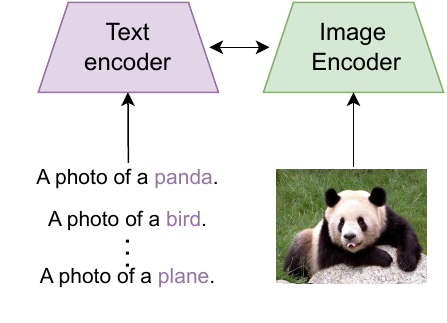}
        \caption{}
        \label{fig:mannual template}
    \end{subfigure}
    \hspace{0.3cm}
    \begin{subfigure}[t]{0.2\linewidth}
		\centering
		\includegraphics[width=\linewidth]{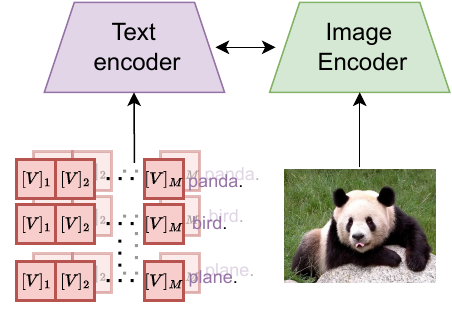}
    \caption{}
		\label{fig:learnable token}
    \end{subfigure}
    \hspace{0.3cm}
    \begin{subfigure}[t]{0.2\linewidth}
		\centering
		\includegraphics[width=\linewidth]{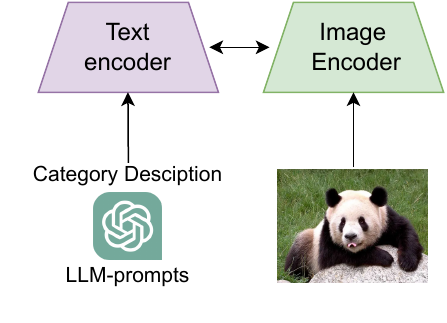}
        \caption{}
		\label{fig:gpt prompt}
	\end{subfigure}
    \hspace{0.3cm}
    \begin{subfigure}[t]{0.3\linewidth}
    \centering
    \includegraphics[width=\linewidth]{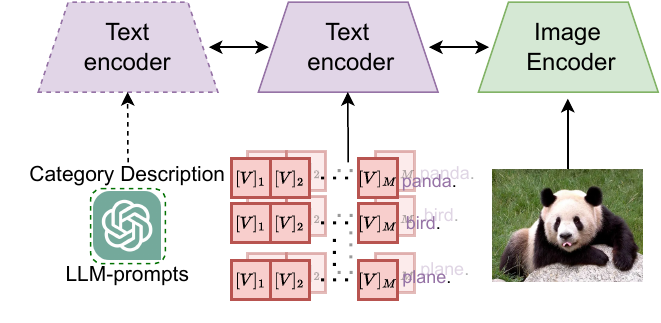}	
    \caption{}
    \label{fig:dual align}
    \end{subfigure}
    \caption{Different paradigms of adapting pre-trained VLM to downstream tasks. (a) Zero-shot inference with manual prompt template. (b) Prompt learning with trainable tokens. (c) Zero-shot inference with class-wise description from LLM. (d) Dual-aligned prompt learning with LLM. Dashed lines indicate that they are only used during the training process.}
\end{figure*}

In this work, we propose the context-aware prompt learning approach for VLMs by effectively transferring knowledge from LLMs.
Instead of directly using the LLMs for inference, we introduce a powerful LLM as a teacher to generate the class-wise descriptions for knowledge distillation.
As is shown in~\Cref{fig:dual align}, the prompts are trained with a dual-alignment strategy that aligns them with both the teacher's knowledge and local image features. 
To focus on nuanced and diverse context information, prompts are explicitly constrained to learn from the verbal descriptions from LLM by a distillation loss.
Additionally, prompts are aligned with local feature tokens through graph matching, which works on both inter- and intra-domain relations.
Only the prompts and the pre-trained VLM are used for downstream task applications such that no additional inference costs are incurred since the external LLM is not used.
This approach not only takes the advantages of both paradigms but also effectively overcomes the  limitations mentioned above.
Compared with previous methods, the prompts can benefit from both local image features and the LLM's knowledge. 
We highlight the following methodological and empirical contributions:

\begin{itemize}
    \item We propose and formalize a prompt learning method, called \textsc{DuAl-PT}, laying down the first work to  bridge the gap between LLM and prompt tuning.
    \item \textsc{DuAl-PT} is a more efficient and effective way to adapt the VLM (like CLIP) to downstream datasets, making use of both learnable prompts for modelling implicit context and diverse explicit descriptions from LLM.
    \item With \textsc{DuAl-PT}, prompt tuning is enhanced via the knowledge of LLM, and the potential of LLM in few-shot domains is  explored. \textsc{DuAl-PT} is also friendly for inference without access to LLM.
    
    \item The proposed method shows superior performance on 11 downstream datasets under few-shot recognition and base-to-new generalization. Further ablation studies and analysis demonstrate the effectiveness of distilling from LLM and graph matching.
\end{itemize}

\section{Methodology}

\subsection{Preliminaries}
\noindent\textbf{Zero-shot inference of CLIP.} 
CLIP~\cite{clip} is a pre-training vision-language model (VLM), consisting of  an image encoder $f(\cdot)$ and a text encoder $g(\cdot)$, and the normalized features of the input image $\mathbf{x}$ and text $\mathbf{t}$ are denoted as $\mathbf{z}$ and $\mathbf{w}$ respectively. To perform zero-shot inference on  $K$ categories, the prompt templates $\{\mathbf{w_i}\}_{i=1}^{K}$ are manually designed  as ``\texttt{A photo of a \{label\}}''~\footnote{For fine-grained classification, it will be ``\texttt{A photo of a \{label\}, a type of \{category\}}'', $e.g.$, ``\texttt{A photo of a \{label\}, a type of flower}'' for Flowers102~\cite{flowers}.}.
The prediction probability for each downstream category is formulated as:
\begin{equation}
    p(y|x) = \frac{\exp{(\cos{(\mathbf{z},\mathbf{w_{y}})}/\tau)}}{\sum_{j=1}^{K}\exp{(\cos{(\mathbf{z},\mathbf{w_{j}})}/\tau)}}
\end{equation}
where $\tau$ and $\cos{(\cdot)}$ are temperature and cosine similarity.

\noindent\textbf{Prompt Learning.} Instead of using pre-defined prompt template, CoOp~\cite{coop}, for the first time, introduces learnable prompts (prompt tuning) to VLMs. 
By replacing the pre-defined template with learnable continuous prompt tokens, downstream class descriptions are obtained via concatenating the prompt tokens with the category name.
The prediction probability is formulated as:
\begin{equation}\label{coop_eq}
    p(y|x) = \frac{\exp{(\cos{(\mathbf{z},\mathbf{w_{y}(S)})}/\tau)}}{\sum_{j=1}^{K}\exp{(\cos{(\mathbf{z},\mathbf{w_{j}(S)})}/\tau)}}
\end{equation}
The prompt $\mathbf{S}$ is optimized by minimizing the cross entropy loss between given label and prompt prediction. 
\subsection{Dual-Aligned Prompt Tuning}

\begin{figure}[!t]
    \centering
    \includegraphics[width={0.482\textwidth}]{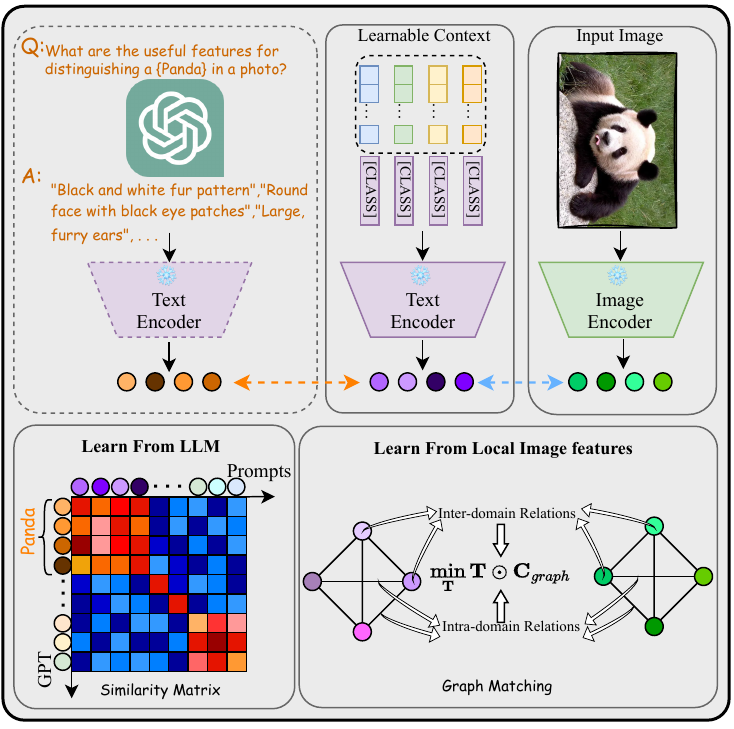}
    \caption{The illustration of DuAL-PT. The prompts are trained in a dual-alignment strategy, with the objective of learning from both the LLM and local images features.}
    \label{fig:main}
\end{figure}

The overall framework of the proposed method is shown in~\cref{fig:main}. The prompts are optimized by learning from auxiliary LLM and local image features. LLM provides explicit context information by generating detailed class-wise descriptions, which are diverse and fine-grained, corresponding to specific local image features.

\subsubsection{Learn From Large Language Model.} As is pre-trained with massive corpora, the large language model has readily learned remarkable world knowledge on a variety of topics, and can be considered as external knowledge bases for downstream datasets. Given $K$ categories, we adopt a unified template to query the LLM: \texttt{Q: What are the useful features for distinguishing a \{CLASS\} in a photo? Please just give me a list of short phrases. Answer: -}~\footnote{``Answer: -" again makes sure the LLM gives a list of descriptions.  }, \\
and the answers $P$ would provide diverse local context information for each class:
\begin{equation}
    H = \text{LLM}(\text{Questions})
\end{equation}
For example, when querying the LLM about \texttt{\{panda\}}, the LLM would generate the answers like ``\texttt{Black and white fur pattern}'', ``\texttt{Round face with black eye patches}'',  ``\texttt{Round body shape with short legs}'', ``\texttt{Distinctive thumb on front paws}'', ``\texttt{Large, furry ears}'', $etc$. Then, for each class $i$, and the context embedding from LLM is formulated as:
\begin{equation}
    \mathbf{h_i} = g(H_i) \in R^{e\times d}
\end{equation}
where $d$ is the dimension of embedding, and $e$ represents the number of context information given by LLM. 

To distill knowledge from LLM, we set multiple ($M$) learnable prompts, denoted as $\mathbf{w_i(S)} \in R ^{M\times d}$, and align the prompts with these explicit descriptions of the corresponding category generated by LLM. 
Cosine distance is applied as a regularization to guide the prompts to be close to their corresponding class-wise descriptions in the embedding space, and the overall distillation loss for all categories is demonstrated as:
\begin{equation}
    L_{\text{LLM}} = \frac{1}{K}\sum^K {1 - \cos (\mathbf{w_i(S)}, \mathbf{h_i})}
\end{equation}

Note that there is also other implementation of the distillation, which is discussed in the ablation studies.

\subsubsection{Learn from local image features.}
We aim to align the multiple prompts with the local image features instead of the global one. The core idea is that each context description generated by LLM only focuses on a specific pattern within a class.
Aligning prompts in this way will encourage the model to concentrate on diverse visual features important for fine-grained classification.
We define it as a graph matching problem, where node and edge matching capture inter- and intra-domain relations, respectively. 
Inter-domain relations facilitate the distribution alignment between the prompts with image features, while the intra-domain relations are motivated by the fact that local image features are highly-related rather than being independent.

Given local image feature vectors $\mathbf{Z}\in \mathcal{R}^{N\times d}$, we construct a graph $G_{\text{img}}(V_{\text{img}},E_{\text{img}})$, with $V_{\text{img}}$ and $E_{\text{img}}$ as nodes and edges.
To fully explore the relations among local features, the graph is fully connected with cosine similarity as edge weights. Similarly, the graph for prompts $G_\text{{text}}(V_{\text{text}},E_{\text{text}})$ is constructed in the same manner. 
We now look into a unified method for aligning the two relations across domain.

\begin{figure}[t!]
\centering
\begin{tcolorbox}[width=1\linewidth] 
    \begin{algorithm}[H]
        \caption{Solve Assignment Matrix.}
        \label{alg}            
        \begin{algorithmic}[1]
            \Require Local feature $\{\mathbf{z}_i\}_{i=1}^N$,prompts $\{\mathbf{w}\}_{i=1}^M$, vector $\mathbf{p},\mathbf{q}$, regularization $\lambda$
        
            \State $\mathbf{C}_{WD} = 1 - \cos(\mathbf{Z},\mathbf{W})$
            \State $\mathbf{C}_{z} = \cos{(\mathbf{Z},\mathbf{Z})}, \mathbf{C}_{w} = \cos{(\mathbf{W},\mathbf{W})}$ 
            \State $\mathbf{C}_{zw} = \mathbf{C}_z^2\mathbf{p}\mathbf{1}_M^\top + \mathbf{1}_N\mathbf{q}(\mathbf{C}_w^2)^\top$ 
            \For{$i=1,2,3,\cdots$}
            \State $\mathbf{C}_{GWD} = \mathbf{C}_{zw} - 2\mathbf{C}_z\mathbf{T}\mathbf{C}_w^\top$ 
            \State $\mathbf{C}_{graph} = \alpha \mathbf{C}_{GWD} + (1-\alpha)\mathbf{C}_{WD}$ 
            \State $\mathbf{a} = \mathbf{1}, \mathbf{K} = \exp{(-\mathbf{C}_{graph}/\lambda)}$ 
            \For{$j=1,2,3,\cdots$}
            \State $\mathbf{a} = \mathbf{p}/\mathbf{Kb}, \mathbf{b} = \mathbf{q}/\mathbf{K}^\top\mathbf{a}$
            \EndFor
            \State $\mathbf{T} = \text{diag}(\mathbf{a})\mathbf{K}\text{diag}(\mathbf{b})$
            \EndFor
        \end{algorithmic}
    \end{algorithm}
\end{tcolorbox}
\end{figure}

We first investigate how prompts are directly corresponding to local visual features. Wasserstein Distance (WD) is a common metric to compare the distributions. Given the cost function $\mathbf{C}_{WD}$ between local visual features and prompts, solving WD in~\cref{eq:wd} learns an assignment plan $\mathbf{T}$ for inter-domain alignment.
\begin{equation}\label{eq:wd}
    D_w(\boldsymbol{Z},\boldsymbol{W}) = \min_{\mathbf{T}} \mathbf{T} \odot \mathbf{C}_{WD} 
\end{equation}
where cosine distance $[\mathbf{C}_{WD}]_{ij} = 1-\cos{(\mathbf{z}_i,\mathbf{w}_j)}$ is set as the cost function in our case, $\odot$ denotes Hadamard product.

\begin{figure*}[t!]
    \centering
    \begin{subfigure}[t]{0.22\linewidth}
	\centering
		\includegraphics[width=\linewidth]{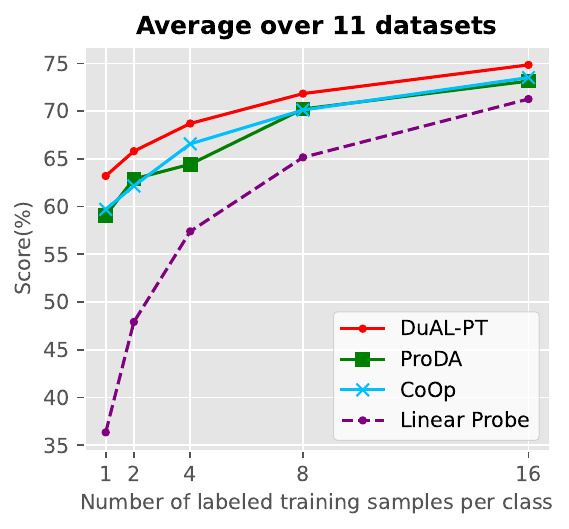}
    \end{subfigure}
    \begin{subfigure}[t]{0.22\linewidth}
		\centering
		\includegraphics[width=\linewidth]{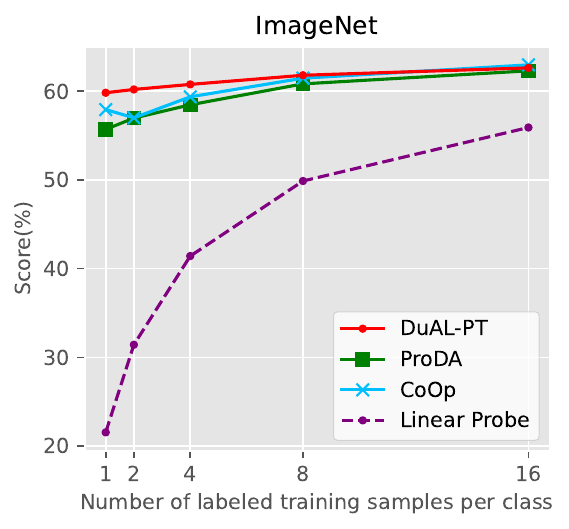}
		\label{ab:dic size}
	\end{subfigure}
    \begin{subfigure}[t]{0.22\linewidth}
		\centering
		\includegraphics[width=\linewidth]{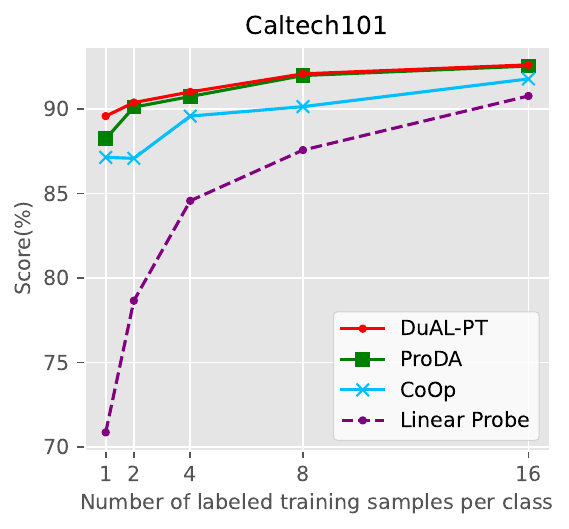}
		\label{ab:dic size}
	\end{subfigure}
    \begin{subfigure}[t]{0.22\linewidth}
		\centering
		\includegraphics[width=\linewidth]{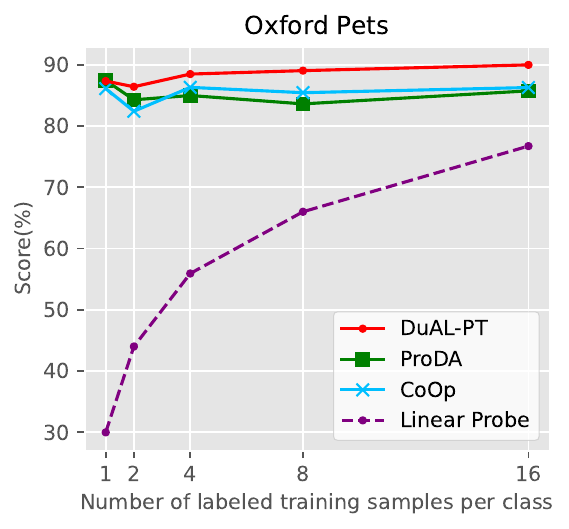}	
		\label{ab:dic size}
	\end{subfigure}
 
    \begin{subfigure}[t]{0.22\linewidth}
		\centering
		\includegraphics[width=\linewidth]{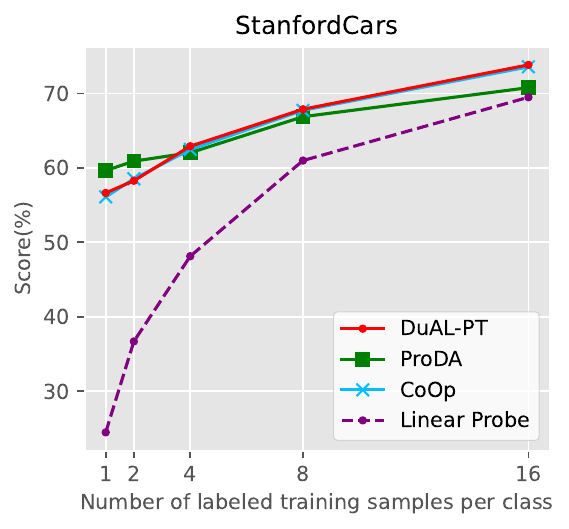}
		\label{ab:dic size}
	\end{subfigure}
    \begin{subfigure}[t]{0.22\linewidth}
		\centering
		\includegraphics[width=\linewidth]{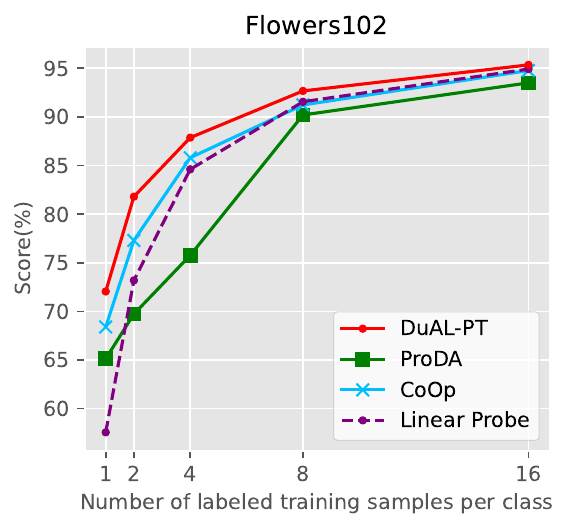}
		\label{ab:dic size}
	\end{subfigure}
    \begin{subfigure}[t]{0.22\linewidth}
		\centering
		\includegraphics[width=\linewidth]{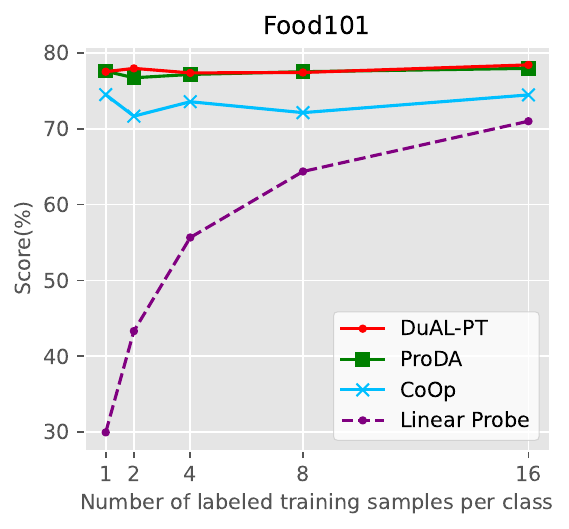}
		\label{ab:dic size}
	\end{subfigure}
    \begin{subfigure}[t]{0.22\linewidth}
		\centering
		\includegraphics[width=\linewidth]{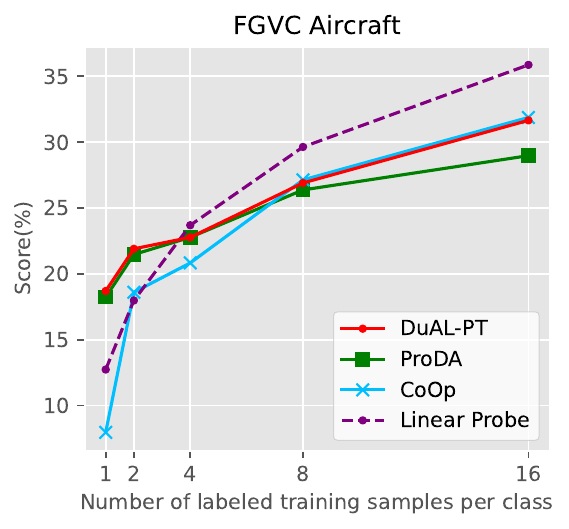}
		\label{ab:dic size}
	\end{subfigure}
 
    \begin{subfigure}[t]{0.22\linewidth}
		\centering
		\includegraphics[width=\linewidth]{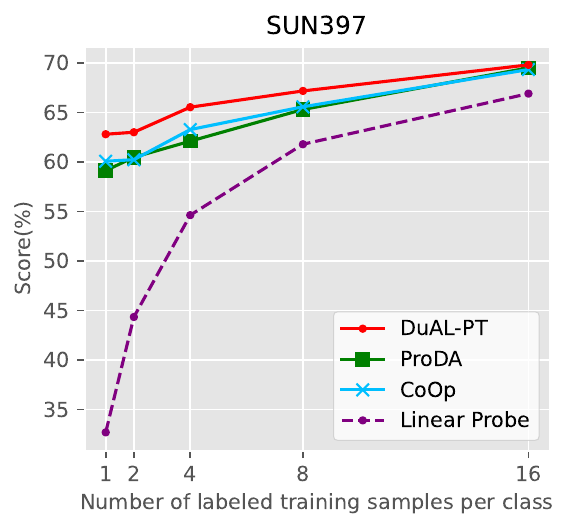}
		\label{ab:dic size}
	\end{subfigure}
    \begin{subfigure}[t]{0.22\linewidth}
		\centering
		\includegraphics[width=\linewidth]{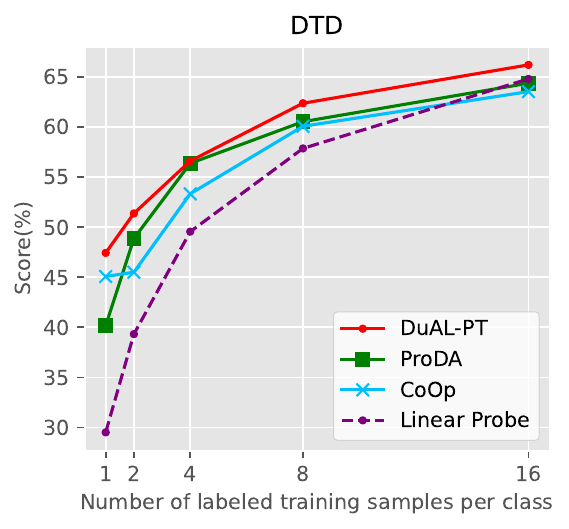}
		\label{ab:dic size}
	\end{subfigure}
    \begin{subfigure}[t]{0.22\linewidth}
		\centering
		\includegraphics[width=\linewidth]{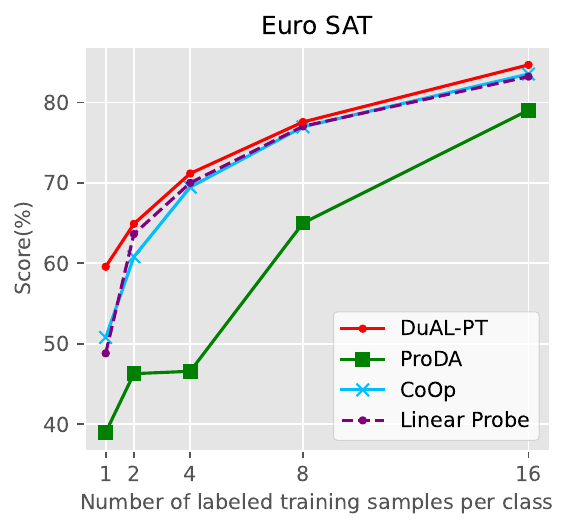}
		\label{ab:dic size}
	\end{subfigure}
    \begin{subfigure}[t]{0.22\linewidth}
		\centering
		\includegraphics[width=\linewidth]{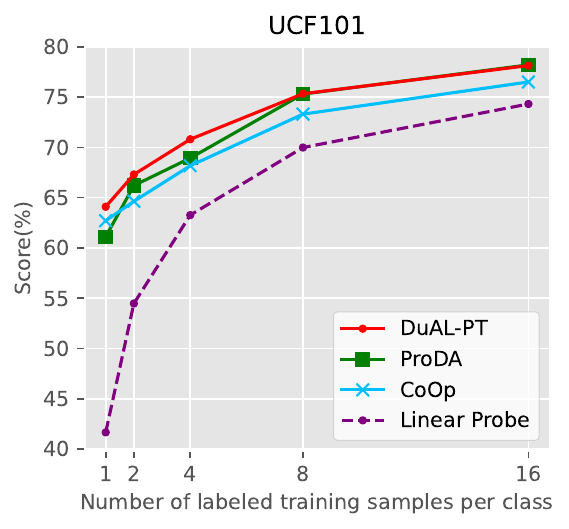}
		\label{ab:dic size}
	\end{subfigure}
    
    \caption{Results of few-shot learning on 11 downstream datasets. The proposed method outperforms others in general.}
    \label{fig:fewshot}
\end{figure*}

As is figured out before, intra-domain relation is also essential for downstream task. We propose to align the edges to maintain this relation. We aim to match the intra-domain relations between local visual features and prompt embedding. It is implemented with Gromov-Wasserstein Distance (GWD), which operates on the distance between pairs of
entities calculated within the domain and measures how these distances compare to those in another domain. Solving GWD gives another plan for aligning two domains for intra-domain relation. Following~\cite{gromov-text,gromov-calculate}, it is defined as:
\begin{align}
    \label{eq:gwd}
    D_{gw}(\boldsymbol{Z},\boldsymbol{W}) &= \min_{\mathbf{T}}  \mathbf{T} \odot \mathbf{C}_{GWD} \nonumber \\
    &= \min_{\mathbf{T}}  \mathbf{T} \odot(\mathbf{C}_{zw} -2\mathbf{C}_z \mathbf{T} \mathbf{C}_w^{\top})  
\end{align}
where $\mathbf{C}_z = \cos(\mathbf{Z},\mathbf{Z}), \mathbf{C}_w = \cos(\mathbf{W},\mathbf{W})$ denote intra-domain similarities and $\mathbf{C}_{zw} = \mathbf{C}_z^2 \mathbf{p}\mathbf{1}_M^{\top} + \mathbf{1}_N\mathbf{C}_w\mathbf{q}(\mathbf{C}_w^2)^{\top}$ represent cross-domain similarities ($\mathbf{p},\mathbf{q}$ are possibility vectors with uniform weights, i.e.,$\mathbf{p}_i=\frac{1}{M}, \mathbf{q}_i=\frac{1}{N}$)


Aligning the two domains with graph matching should consider these two relations at the same time. Therefore, we combine $\mathbf{C}_{WD}$ and $ \mathbf{C}_{GWD}$ with a weighting parameter $\alpha$ as an overall, and the final assignment matrix is solved with Sinkhorn-Knopp algorithm~\cite{sinkhorn}. The complete solution to graph matching is shown in Algorithm~\ref{alg}.


\subsubsection{Training Objective.}
The obtained assignment matrix $\mathbf{T}$ shows how the local visual features and prompts are related, which is further interpreted as the weights to ensemble various prompts. We formulate the prediction probability in Eq.\ref{eq:p}:
\begin{equation}\label{eq:p}
    p(y=k|x) = \frac{\exp( \mathbf{T}\odot (\mathbf{z}\mathbf{w}(k))/\tau)}{\sum_{k'} \exp ( \mathbf{T}\odot (\mathbf{z} \mathbf{w}(k^{'}))/\tau)}
\end{equation}
Implicit context information is learned by applying cross-entropy loss $L_\text{img}$ with $p(y=k|x)$ and the label.

To learn from LLM and local image features for more diverse context, the overall loss function is demonstrated in Eq.\ref{all_loss}, with $\beta$ to balance the dual alignment:
\begin{equation} \label{all_loss}
    L = \beta L_{\text{LLM}} + (1-\beta) L_{\text{img}}
\end{equation}
\subsubsection{Inference.} Only the prediction $p(y=k|x)$ will be counted for downstream inference. Without further access to external LLM, the proposed method is friendly to deployment, and can be easily adapted to unseen categories.

\begin{table*}[!ht]

\centering
\caption{Comparison of CoCoOp and Co\textsc{DuAl-PT} (conditional \textsc{DuAl-PT}) in base-to-new generalization. \textbf{H}: Harmonic mean.
}
\label{tab:base2new}
    \begin{subtable}[t]{0.24\linewidth}
    \centering
    \caption{\textbf{Average.}}
    \resizebox{!}{0.62cm}{
    \begin{tabular}{ccc|c}
    \toprule
           & Base & New & H \\
    \midrule
    CoCoOp & 75.10     &  63.73   & 67.84  \\
    \rowcolor[HTML]{EFEFEF}
    Co\textsc{DuAl-PT} & \textbf{75.50 }    &  \textbf{66.09}   &\textbf{69.95}   \\
    \bottomrule
    \end{tabular}
    }
    \end{subtable}
    \begin{subtable}[t]{0.24\linewidth}
    \centering
    \caption{ImageNet.}
    \resizebox{!}{0.62cm}{
    \begin{tabular}{ccc|c}
    \toprule
           & Base & New & H \\
    \midrule
    CoCoOp & 68.30    & \textbf{62.80}    &\textbf{65.43}  \\
    \rowcolor[HTML]{EFEFEF}
    Co\textsc{DuAl-PT} &  \textbf{68.42}    & 62.17    & 65.15  \\
    \bottomrule
    \end{tabular}
    }
    \end{subtable}
    \begin{subtable}[t]{0.24\linewidth}
    \centering
    \caption{Caltech101}
    \resizebox{!}{0.62cm}{
    \begin{tabular}{ccc|c}
    \toprule
           & Base & New & H \\
    \midrule
    CoCoOp & 94.35     & 89.77   &92.00\\
    \rowcolor[HTML]{EFEFEF}
    Co\textsc{DuAl-PT} &  \textbf{95.00 }   & \textbf{90.17}    & \textbf{92.52} \\
    \bottomrule
    \end{tabular}
    }
    \end{subtable}
    \begin{subtable}[t]{0.24\linewidth}
    \centering
    \caption{OxfordPets}
    \resizebox{!}{0.62cm}{
    \begin{tabular}{ccc|c}
    \toprule
           & Base & New & H \\
    \midrule
    CoCoOp &92.40      & \textbf{96.07}    & \textbf{94.19}  \\
    \rowcolor[HTML]{EFEFEF}
    Co\textsc{DuAl-PT} & \textbf{92.60}      & 94.80    & 93.69  \\
    \bottomrule
    \end{tabular}
    }
    \end{subtable}
    \vspace{0.5em}

    \begin{subtable}[t]{0.24\linewidth}
    \centering
    \caption{StanfordCars.}
    \resizebox{!}{0.62cm}{
    \begin{tabular}{ccc|c}
    \toprule
           & Base & New & H \\
    \midrule
    CoCoOp &63.63      & 64.50    &64.06\\
    \rowcolor[HTML]{EFEFEF}
    Co\textsc{DuAl-PT} & 63.63     & \textbf{65.46}    &\textbf{64.53}   \\
    \bottomrule
    \end{tabular}
    }
    \end{subtable}
    \begin{subtable}[t]{0.24\linewidth}
    \centering
    \caption{Flowers102.}
    \resizebox{!}{0.62cm}{
    \begin{tabular}{ccc|c}
    \toprule
           & Base & New & H \\
    \midrule
    CoCoOp & 89.43  & 67.60    & 76.99  \\
    \rowcolor[HTML]{EFEFEF}
    Co\textsc{DuAl-PT} & \textbf{89.50}     &\textbf{67.77}      & \textbf{77.13}  \\
    \bottomrule
    \end{tabular}
    }
    \end{subtable}
    \begin{subtable}[t]{0.24\linewidth}
    \centering
    \caption{Food101.}
    \resizebox{!}{0.62cm}{
    \begin{tabular}{ccc|c}
    \toprule
           & Base & New & H \\
    \midrule
    CoCoOp & 83.40  & 83.60    & 83.50   \\
    \rowcolor[HTML]{EFEFEF}
    Co\textsc{DuAl-PT} & \textbf{83.83}     &   \textbf{84.40}  & \textbf{84.11}  \\
    \bottomrule
    \end{tabular}
    }
    \end{subtable}
    \begin{subtable}[t]{0.24\linewidth}
    \centering
    \caption{FGVCAircraft.}
    \resizebox{!}{0.62cm}{
    \begin{tabular}{ccc|c}
    \toprule
           & Base & New & H \\
    \midrule
    CoCoOp & 22.87     & 16.17    & 18.95  \\
    \rowcolor[HTML]{EFEFEF}
    Co\textsc{DuAl-PT} & \textbf{24.07}     &  \textbf{21.83}   &\textbf{22.89}\\
    \bottomrule
    \end{tabular}
    }
    \end{subtable}
    \vspace{0.5em}

    \begin{subtable}[t]{0.24\linewidth}
    \centering
    \caption{SUN397.}
    \resizebox{!}{0.62cm}{
    \begin{tabular}{ccc|c}
    \toprule
           & Base & New & H \\
    \midrule
    CoCoOp & 74.37     & 73.40    & 73.88   \\
    \rowcolor[HTML]{EFEFEF}
    Co\textsc{DuAl-PT} & \textbf{74.47}     & \textbf{73.70}    & \textbf{74.08}  \\
    \bottomrule
    \end{tabular}
    }
    \end{subtable}
    \begin{subtable}[t]{0.24\linewidth}
    \centering
    \caption{DTD.}
    \resizebox{!}{0.62cm}{
    \begin{tabular}{ccc|c}
    \toprule
           & Base & New & H \\
    \midrule
    CoCoOp &  71.83    &  46.37   &56.35  \\
    \rowcolor[HTML]{EFEFEF}
    Co\textsc{DuAl-PT} & \textbf{73.13}     &   \textbf{49.13}  &\textbf{58.77 } \\
    \bottomrule
    \end{tabular}
    }
    \end{subtable}
    \begin{subtable}[t]{0.24\linewidth}
    \centering
    \caption{EuroSAT.}
    \resizebox{!}{0.62cm}{
    \begin{tabular}{ccc|c}
    \toprule
           & Base & New & H \\
    \midrule
    CoCoOp &  88.50    & 34.47    & 49.62  \\
    \rowcolor[HTML]{EFEFEF}
    Co\textsc{DuAl-PT} &  \textbf{88.70}    & \textbf{51.97}    & \textbf{65.54}  \\
    \bottomrule
    \end{tabular}
    }
    \end{subtable}
    \begin{subtable}[t]{0.24\linewidth}
    \centering
    \caption{UCF101.}
    \resizebox{!}{0.62cm}{
    \begin{tabular}{ccc|c}
    \toprule
           & Base & New & H \\
    \midrule
    CoCoOp &  77.00    & 66.30    & 71.25  \\
    \rowcolor[HTML]{EFEFEF}
    Co\textsc{DuAl-PT} & \textbf{77.10}    & \textbf{67.40}    & \textbf{71.92}  \\
    \bottomrule
    \end{tabular}
    }
    \end{subtable}

\end{table*} 
\section{Experiments}
\noindent\textbf{Datasets.} Following CoOp~\cite{coop}, we evaluate the proposed method on 11 downstream datasets, including ImageNet~\cite{imagenet}, Caltech101~\cite{caltech101}, OxfordPets~\cite{pets}, StanfordCars~\cite{cars}, Flowers102~\cite{flowers}, Food101~\cite{food}, FGVCAircraft~\cite{aircraft}, SUN397~\cite{sun}, DTD~\cite{dtd}, EuroSAT~\cite{eurosat}, and UCF101~\cite{ucf101}. Dataset details are provided in the supplementary materials.

\noindent\textbf{Evaluation Protocol.} We evaluate the proposed method under few-shot learning and base-to-new generalization. For few-shot learning, the prompt is trained on 1,2,4,8,16 shots in each class, and tests on the complete validation set. For base-to-new generalization, the prompt is trained with 16 samples per category on base classes and tested on new classes. We report the average results over 3 random seeds.

\noindent\textbf{Baseline Methods.} For few-shot learning, we compare \textsc{DuAl-PT} with CoOp~\cite{coop}, ProDA~\cite{proda}, and linear probe CLIP~\cite{clip}. CoOp is pioneering in prompt learning, while ProDA learns diverse prompts from distributions. These methods are competitive in our community. For base-to-new generalization, CoCoOp~\cite{cocoop} is a strong baseline work.

\noindent\textbf{Implementation Details.} Following the widely-used experimental settings in CoOp, ResNet50~\cite{resnet} pre-trained with CLIP is selected as the image encoder. The local feature map obtained from the last pooling layer has a size of $7\times 7$. The number of local prompts is set as 4 in our experiments. 
We adopt GPT-3.5~\cite{gpt} with a temperature of 0.7 to generate local visual context. The scaling factor $\alpha$ in graph matching and the weight $\beta$ for $L_{LLM}$ are both 0.2. We train the prompts with SGD optimizer with an initial learning rate of 0.002, which decays by the cosine annealing rule. The maximum epoch is 50 for 1 shot, 100 for 2/4 shots, 200 for 8/16 shots, and 10 for base-to-new generalization. Note that \textsc{DuAl-PT} has 4 prompts rather than 1, we improve the length of learnable tokens from 16 to 64 in CoOp to reach the same number of trainable parameters for fair comparison.

\subsection{Few-Shot Learning}
\Cref{fig:fewshot} shows the comparison with the baseline methods on 11 downstream datasets. We notice that \textsc{DuAl-PT} surpasses CoOp (blue) on most datasets, validating the idea of learning enriched context from LLM and local image features. Moreover, compared with ProDA (green), our method is more robust on downstream datasets.
Also working on diverse prompts for more context, ProDA fails to achieve promising results on datasets like Flowers102 and EuroSAT, while the proposed method is capable of handling such a fine-grained task, which demonstrates that \textsc{DuAl-PT} manages to learn evident context from the LLM. In general, the proposed method surpasses the baselines by 3.49\%, 3.60\%, 2.10\%, 1.73\%, and 1.33\% on 1/2/4//8/16 shots in average over all datasets. 

\subsection{Base-to-New Generalization}
We further apply \textsc{DuAl-PT} to a conditional setting to evaluate how well the proposed method can generalize to new classes. For implementation, following CoCoOp, we build a meta net to generate for each input a conditional token, which is then combined with the context vector as an image-specific prompt. Table~\ref{tab:base2new} shows that Co\textsc{DuAl-PT} outperforms CoCoOp on most datasets, especially on FGVCAircraft, DTD and EuroSAT, where fine-grained context is  important. Generally, \textsc{DuAl-PT} improves the baseline by 0.40\% , 2.36\%, 2.11\% in base, new and Harmonic. 
Moreover, without access to LLM when inference on new classes, the proposed \textsc{DuAl-PT} outperforms the baseline by a large margin, further demonstrating that the prompts have learned broad class-wise context for generalization.

\begin{figure*}[t]
    \centering
    \includegraphics[width=0.85\textwidth]{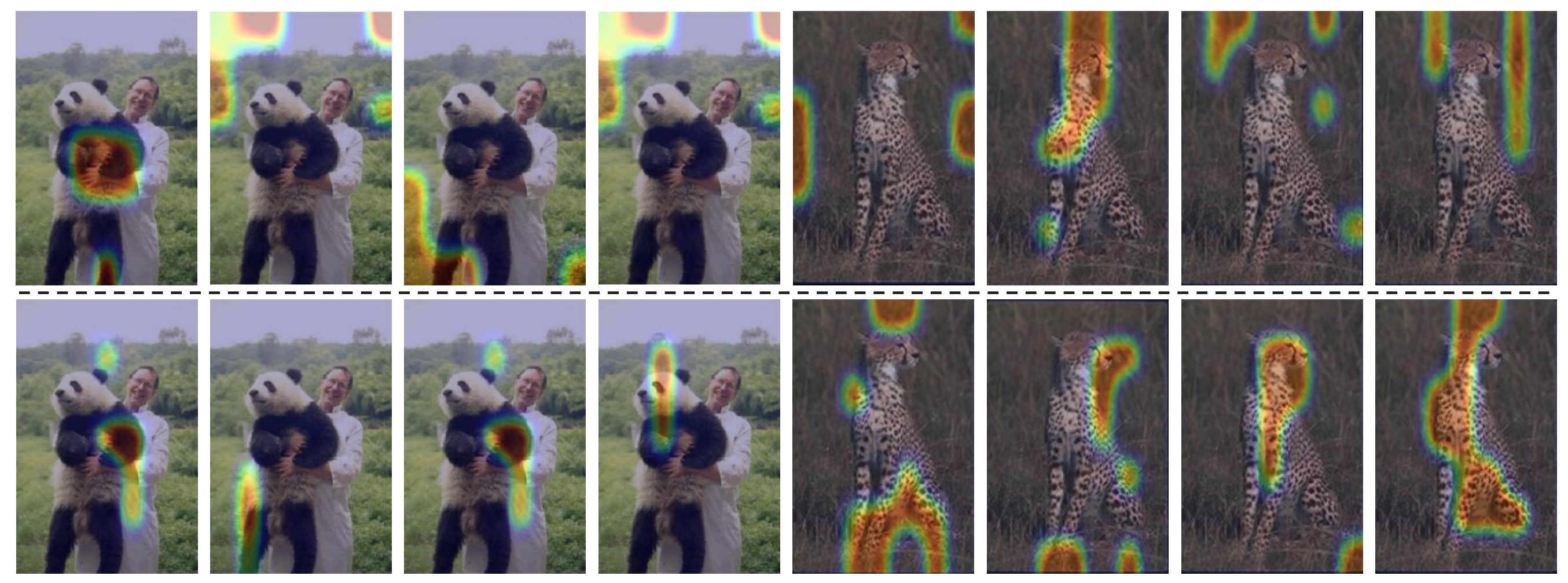}
    \caption{Visualization of the learned prompts. Aligned \textbf{without} (1st row) v.s. \textbf{with} LLM (2nd row) .}
    \label{fig:vis}
\end{figure*}

\subsection{Ablation Study and Further Discussion} \label{sec:ablation}
We conduct more ablation studies on Caltech101, DTD. These two datasets cover general and fine-grained recognition tasks, thus are representative to investigate the effectinveness the components.

\noindent\textbf{How to distill from LLM?} As LLM provides enriched context information beneficial for visual recognition, learning from LLM is of great significance.
\begin{center}{
\resizebox{0.8\linewidth}{!}
{
\begin{tabular}{ccccccc}
Dataset                  & Loss       & Shot1          & Shot2          & Shot4         & Shot8          & Shot16         \\ \hline
\multirow{3}{*}{Caltech} & WD         & 89.50          & 90.27          & 90.87         & \textbf{91.63} & \textbf{92.50}          \\                       
                         & CE         & 87.37          & 89.27          & 90.27         & 90.27          & 91.00              \\
                         & Cosine     & \textbf{89.57} & \textbf{90.37} & \textbf{91.00}   & 91.33       & 92.43 \\ \hline
\multirow{3}{*}{DTD}     & WD         &47.03           &51.37           &54.83          &62.27           &\textbf{66.37}             \\
                         & CE         &45.23	          &49.63	       &53.83	        &60.00	            &63.70                \\
                         & Cosine     & \textbf{47.43} & \textbf{51.37} & \textbf{56.60} & \textbf{62.37} & 66.20  \\ 
\end{tabular}}}
\end{center}
We investigate Wasserstein Distance (WD), instance-wise cross-entropy (CE) and the adopted cosine distance (Cosine).
WD and Cosine both work on class-wise distillation, dependent of input samples. Such a soft regularization achieves comparable but promising performance. CE is an instance-wise supervision, however, falls behind from the other two. We infer that instance-wise logits are noisy for learning class-wise knowledge. 
We adopt cosine distance by default, considering the computational cost of WD.

\noindent\textbf{Other alternatives to learn context information?} LLM provides external context from a language's perspective. Table.~\ref{table:ab_main} demonstrates other two methods to learn context. \textbf{CAM} (Class Activation Map) explicitly highlights the regions of interest, thus offers more context information from visual domain. \textbf{Sim} calculates the importance of local feature tokens with the global token, and only top 50\% tokens are selected to contribute to final classification result. This strategy filters out irrelevant regions, giving local context from a feature's perspective. Experimental results show that learning from LLM achieves the best performance. We infer that \textbf{CAM} gives instance-wise context, which is not general within the class, and similarity with global representation (\textbf{Sim}) cannot explicitly represent fine-grained variance.

\noindent\textbf{How does LLM benefit prompt learning?} 
As is shown in Table~\ref{table:ab_main}, LLM improves few-shot recognition by 1.20\%, 2.80\%, 1.80\%, 2.13\%, and 0.06\% on Caltech, 1.00\%, 1.10\%, 0.67\%, 1.07\%, and 1.40\% on DTD (GPT + Graph vs None + Graph). 
Fig.~\ref{fig:vis} further shows that, with the guidance of LLM, prompts are learned to focus more on the key visual features, demonstrating the effectiveness of the LLM.

\begin{table}[t!]
\caption{Ablation Studies on different implementation and other alternatives of the components.}
\resizebox{\linewidth}{!}{
\begin{tabular}{cccccccc}
\hline
Dataset                      & Context                     & Align                         & Shot1                                  & Shot2                                  & Shot4                                  & Shot8                                  & Shot16                                 \\ \hline
                             & \cellcolor[HTML]{FFFFC7}None                        & \cellcolor[HTML]{FFFFC7}Graph                         & \cellcolor[HTML]{FFFFC7}88.37                                  & \cellcolor[HTML]{FFFFC7}87.57                                  & \cellcolor[HTML]{FFFFC7}89.20                                  & \cellcolor[HTML]{FFFFC7}89.20                                  & \cellcolor[HTML]{FFFFC7}92.37                                   \\
                             & \cellcolor[HTML]{CBCEFB}CAM                         & \cellcolor[HTML]{CBCEFB}Graph                         & \cellcolor[HTML]{CBCEFB}88.90                                  & \cellcolor[HTML]{CBCEFB}87.67                                  & \cellcolor[HTML]{CBCEFB}89.43                                  & \cellcolor[HTML]{CBCEFB}90.70                                  & \cellcolor[HTML]{CBCEFB}\textbf{92.57}                         \\
                             & \cellcolor[HTML]{CBCEFB}SIM                         & \cellcolor[HTML]{CBCEFB}Graph                         & \cellcolor[HTML]{CBCEFB}88.57                                  &\cellcolor[HTML]{CBCEFB} 87.73                                  & \cellcolor[HTML]{CBCEFB}89.40                                  &\cellcolor[HTML]{CBCEFB} 90.67                                  & \cellcolor[HTML]{CBCEFB}92.43                                  \\
                             & \cellcolor[HTML]{D9F5D6}GPT                         & \cellcolor[HTML]{D9F5D6}Attn                          &\cellcolor[HTML]{D9F5D6} 87.73                                  & \cellcolor[HTML]{D9F5D6}88.30                                  & \cellcolor[HTML]{D9F5D6}89.47                                  & \cellcolor[HTML]{D9F5D6}89.63                                  & \cellcolor[HTML]{D9F5D6}91.83                                  \\
                             & \cellcolor[HTML]{D9F5D6}GPT                         & \cellcolor[HTML]{D9F5D6}Node                          &\cellcolor[HTML]{D9F5D6} 89.37                                  & \cellcolor[HTML]{D9F5D6}90.00                                  & \cellcolor[HTML]{D9F5D6}90.10                                  & \cellcolor[HTML]{D9F5D6}90.67                                  & \cellcolor[HTML]{D9F5D6}92.43                                  \\
                             &\cellcolor[HTML]{D9F5D6} GPT                         & \cellcolor[HTML]{D9F5D6}Edge                          &\cellcolor[HTML]{D9F5D6} 88.90                                  & \cellcolor[HTML]{D9F5D6}89.80                                  & \cellcolor[HTML]{D9F5D6}90.27                                  & \cellcolor[HTML]{D9F5D6}91.27                                  &\cellcolor[HTML]{D9F5D6} 91.83                                  \\
\multirow{-7}{*}{{\rotatebox{90}{Caltech101}}} & \cellcolor[HTML]{DEE0E3}GPT & \cellcolor[HTML]{DEE0E3}Graph & \cellcolor[HTML]{DEE0E3}\textbf{89.57} & \cellcolor[HTML]{DEE0E3}\textbf{90.37} & \cellcolor[HTML]{DEE0E3}\textbf{91.00} & \cellcolor[HTML]{DEE0E3}\textbf{91.33} & \cellcolor[HTML]{DEE0E3}92.43          \\ \hline
                             & \cellcolor[HTML]{FFFFC7}None                        & \cellcolor[HTML]{FFFFC7}Graph                         & \cellcolor[HTML]{FFFFC7}46.43                                  & \cellcolor[HTML]{FFFFC7}50.27                                  & \cellcolor[HTML]{FFFFC7}55.93                                  & \cellcolor[HTML]{FFFFC7}61.30                                  &\cellcolor[HTML]{FFFFC7} 64.80                                  \\
                             & \cellcolor[HTML]{CBCEFB}CAM                         & \cellcolor[HTML]{CBCEFB}Graph                         &\cellcolor[HTML]{CBCEFB} 47.00                                  & \cellcolor[HTML]{CBCEFB}51.03                                  & \cellcolor[HTML]{CBCEFB}56.50                                  & \cellcolor[HTML]{CBCEFB}61.47                                  & \cellcolor[HTML]{CBCEFB}65.10                                  \\
                             & \cellcolor[HTML]{CBCEFB}SIM                         & \cellcolor[HTML]{CBCEFB}Graph                         & \cellcolor[HTML]{CBCEFB}47.03                                  & \cellcolor[HTML]{CBCEFB}50.20                                  & \cellcolor[HTML]{CBCEFB}56.50                                  & \cellcolor[HTML]{CBCEFB}60.60                                  &\cellcolor[HTML]{CBCEFB} 64.57                                  \\
                             & \cellcolor[HTML]{D9F5D6}GPT                         & \cellcolor[HTML]{D9F5D6}Attn                          & \cellcolor[HTML]{D9F5D6}43.33                                  & \cellcolor[HTML]{D9F5D6}46.40                                  & \cellcolor[HTML]{D9F5D6}53.13                                  & \cellcolor[HTML]{D9F5D6}59.67                                  &\cellcolor[HTML]{D9F5D6} 62.83                                  \\
                             & \cellcolor[HTML]{D9F5D6}GPT                         & \cellcolor[HTML]{D9F5D6}Node                          & \cellcolor[HTML]{D9F5D6}46.83                                  & \cellcolor[HTML]{D9F5D6}\textbf{51.53}                         & \cellcolor[HTML]{D9F5D6}54.63                                  & \cellcolor[HTML]{D9F5D6}61.30                                  & \cellcolor[HTML]{D9F5D6}65.23                                  \\
                             & \cellcolor[HTML]{D9F5D6}GPT                         & \cellcolor[HTML]{D9F5D6}Edge                          & \cellcolor[HTML]{D9F5D6}47.03                                  & \cellcolor[HTML]{D9F5D6}51.10                                  & \cellcolor[HTML]{D9F5D6}54.57                                  &\cellcolor[HTML]{D9F5D6} 61.80                                  & \cellcolor[HTML]{D9F5D6}65.40                                  \\
\multirow{-7}{*}{\rotatebox{90}{DTD}}        & \cellcolor[HTML]{DEE0E3}GPT & \cellcolor[HTML]{DEE0E3}Graph & \cellcolor[HTML]{DEE0E3}\textbf{47.43} & \cellcolor[HTML]{DEE0E3}51.37          & \cellcolor[HTML]{DEE0E3}\textbf{56.60} & \cellcolor[HTML]{DEE0E3}\textbf{62.37} & \cellcolor[HTML]{DEE0E3}\textbf{66.20} \\ \hline
\end{tabular}
}
    
\label{table:ab_main}
\end{table}
\noindent\textbf{How to align multiple prompts?} Considering both inter- and intra- domain relationships, we construct a graph in each domain and align the domains by matching
graphs. Applying cross-attention (\textbf{Attn}) to generate a dense mapping function is a simple and direct idea. However, it is shown in Table~\ref{table:ab_main} that cross-attention alignment
achieves the poorest performance. Essentially, it only obtains the alignment relations with feature similarity, ignoring more comprehensive relations. Moreover, capturing inter- and intra-relations is mutually beneficial to few-shot recognition, demonstrating that it is practical to align prompts and local features with graph. (\textbf{Graph vs Node/Edge}).

\begin{figure}[t]
    \centering
    \includegraphics[width=0.7\linewidth]{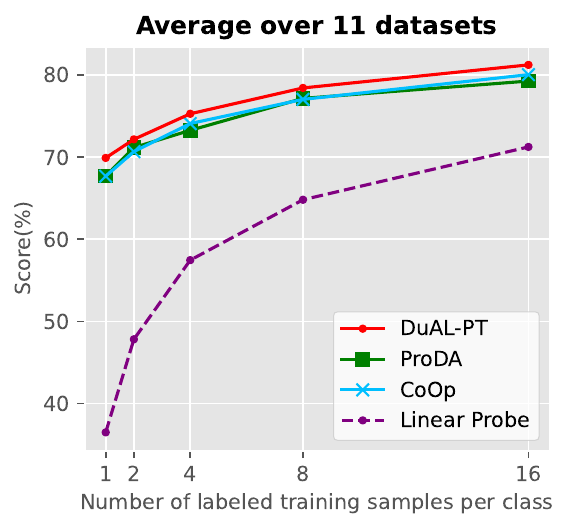}
    \caption{Few-shot recognition on 11 datasets with ViT. }
    \label{vit}
\end{figure}

\noindent\textbf{How does \textsc{DuAl-PT} work on ViT?} More experiments are conducted on CLIP-ViT-B/16 backbone. Fig.\ref{vit} shows that \textsc{DuAl-PT} outperforms other methods over 2.22\%, 0.99\%, 2.02\%, 1.27\%, 1.97\% on 1/2/4/8/16 shots on average over all datasets. Working on different backbones, \textsc{DuAl-PT} can consistently outperform other methods, demonstrating the effectiveness and generalization ability of \textsc{DuAl-PT}.

\begin{figure*}[t]
    \centering
    \begin{minipage}{0.24\linewidth}
		\centering
		\includegraphics[width=\linewidth]{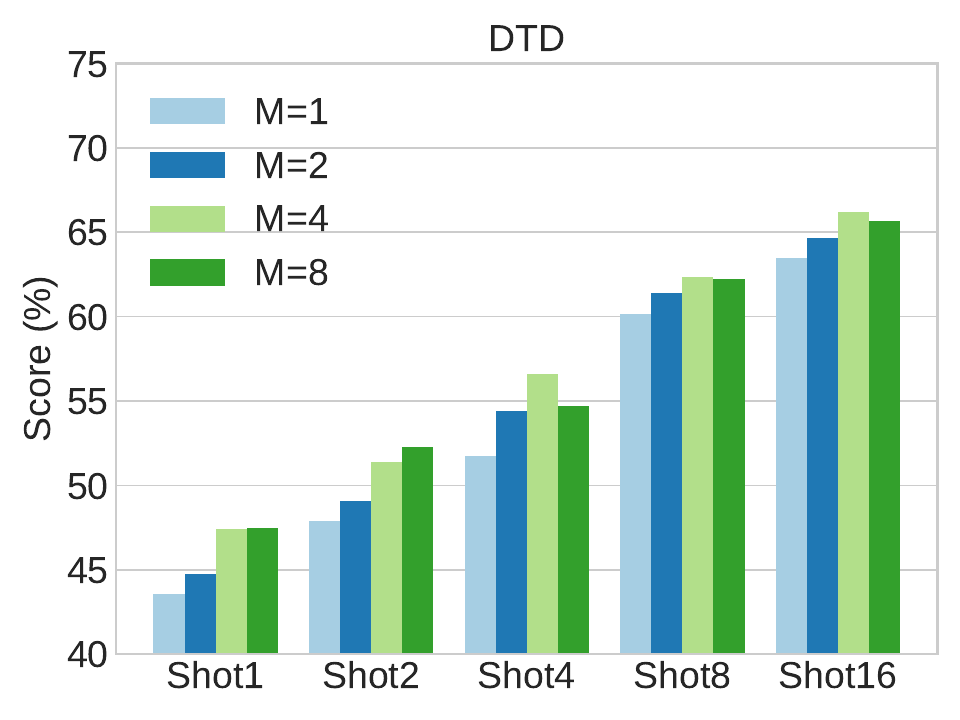}
		
            \subcaption{}
            \label{ab:m1}
	\end{minipage}
     \begin{minipage}{0.24\linewidth}
		\centering
		\includegraphics[width=\linewidth]{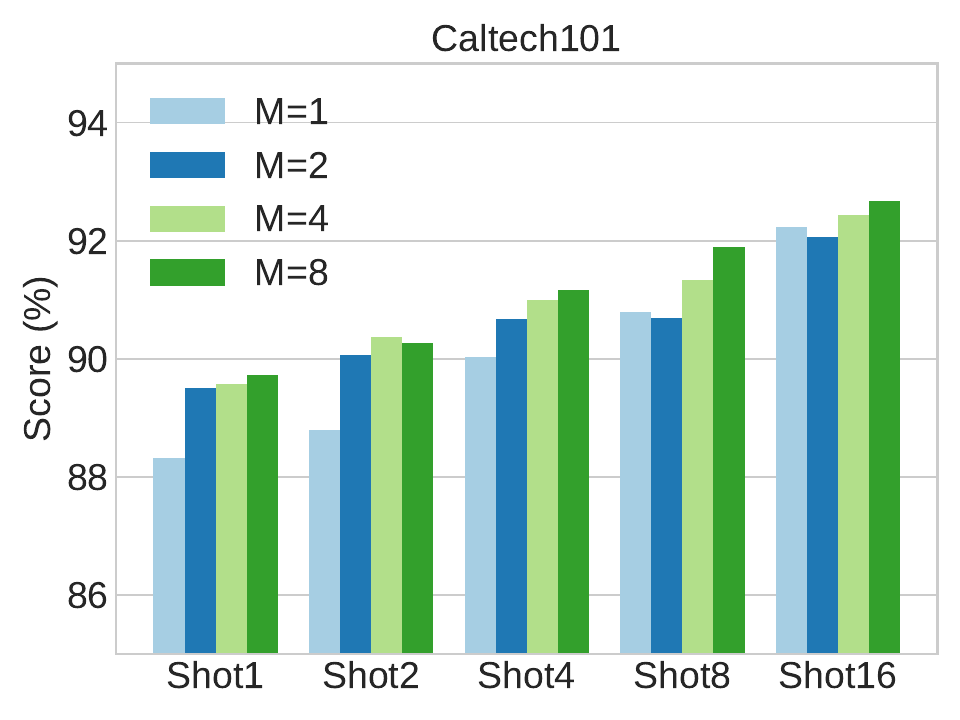}
		
            \subcaption{}
            \label{ab:m2}
	\end{minipage}
    \begin{minipage}{0.24\linewidth}
		\centering
		\includegraphics[width=\linewidth]{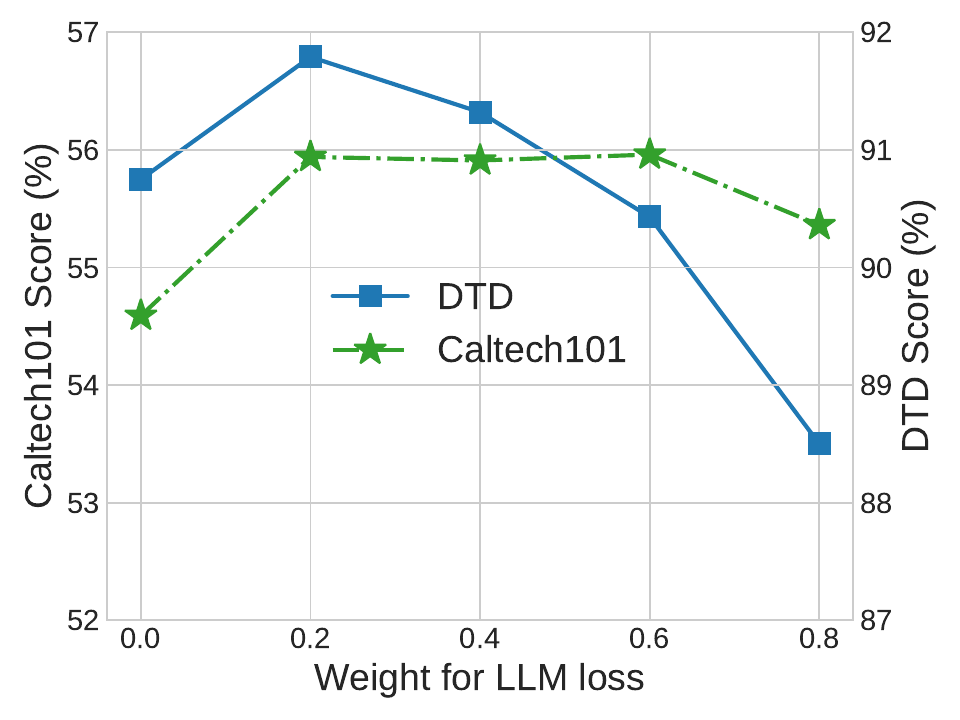}
		
            \subcaption{}
            \label{ab:loss}
	\end{minipage}
    \begin{minipage}{0.24\linewidth}
		\centering
		\includegraphics[width=\linewidth]{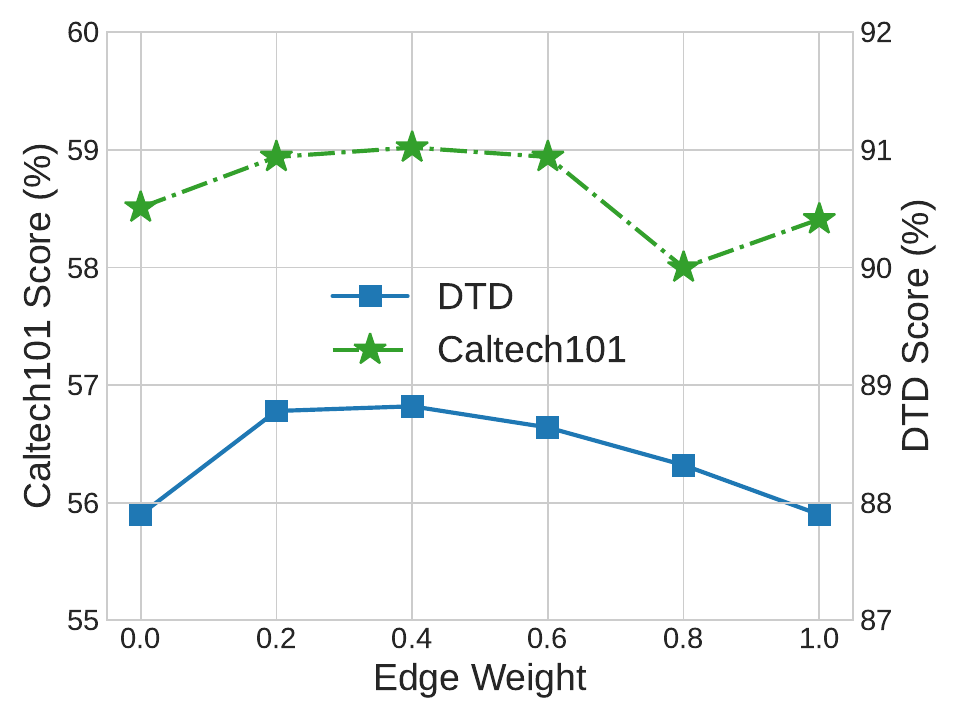}
		
        \subcaption{}
        \label{ab:graph}
	\end{minipage}
    \caption{Ablation study on the hyper-parameters. (a)(b) Number of local prompts. (c) Weight for LLM loss in the dual-alignment strategy. (Average of 1/2/4/8/16 shots) (d) Edge weight in graph match. 
 (Average of 1/2/4/8/16 shots)}
\end{figure*}

\noindent\textbf{Number of local prompts.} We study the number $M$ of local prompts to align with the local image features and LLM descriptions under the setting of few-shot learning. ~\cref{ab:m1,ab:m2} show that more local prompts are potentially beneficial, while there is no obvious improvement when $M>4$ on DTD. Considering computation and time consumption, we set $M$ as 4 in main experiments as a trade-off solution.

\noindent\textbf{Balancing learning from LLM and local image features.} The prompts are trained with a dual-alignment strategy, and we further investigate the scaling factor to balance different branches. ~\cref{ab:loss} demonstrates that distilling from LLM benefits downstream classification, and soft regularization for this branch with less weight achieves better results.

\noindent\textbf{Balancing edge and node in graph match.} we match two graphs by combining the cost function of WD and GWD. We ablate the weight $\alpha$ of the GWD (edge) cost function. The results in~\cref{ab:graph} reveal that setting $\alpha$ as 0.2 achieves promising results in both datasets. Notably, if two graphs are matched by the edge or node alone ($\alpha= 0/1$), the overall performance drops, indicating inter- and intra- relationships are mutually beneficial for learning good prompts.

\noindent\textbf{Analysis of inference time.} Compared with CoOp, \textsc{DuAl-PT} requires 10\% extra time. It mainly comes from graph matching, which is of $O(N^4)$ time complexity (N=4 in \textsc{DuAl-PT}). Given the gain, little extra time is acceptble.

\section{Related Works}
\subsection{Vision-Language Pre-training}
Vision-language models are capable of learning broad and generic visual concepts, showing promising generalization ability to a large variety of downstream tasks and datasets. With the objective of learning the connection between visual content and language, a large number of early works aim to learn representations by predicting the captions of images~\cite{virtex,ngram,radford2019,learn_caption}. However, the main obstacle is that these models are trained on relatively small datasets, e.g., Flickr~\cite{flickr} and COCO Captions~\cite{virtex}. Recent vision-language models based on contrastive learning have shown superb results, benefiting from web-scale image-text pairs. In particular, CLIP~\cite{clip} and ALIGN~\cite{align}, aim to learn the aligned representations of image and text by contrastive loss. Later representative works like ALBEF~\cite{albef} and BLIP~\cite{blip} integrate contrastive loss with other pretext tasks for better pre-training and empower the transferability of VLM.


\subsection{Prompt Learning}

Inspired by the success of \emph{prompt} in the NLP domain, numerous studies introduce prompt learning to better adapt the pre-trained CLIP to downstream datasets, and the core idea is to align learnable prompts with image representations. CoOp~\cite{coop} is the pioneer work in this field. It optimizes continuous prompts by minimizing the classification loss on target datasets, and achieves promising results in adapting pre-trained models to downstream few-shot recognition task. Based on CoOp, more recent works like CoCoOp~\cite{cocoop} and ProDA~\cite{proda} aim at better generalization by aligning visual representation with image-specific prompts or the distribution of diverse prompts. However, the learned context is not always useful only by a standard classification loss with the category name, which motivates us to query the powerful large language model to access abundant knowledge.

\subsection{Large Language Model for Image Classification}
LLMs like GPT-3~\cite{gpt3}, OPT~\cite{opt}, and PaLM~\cite{palm} are trained on massive web-scale corpora, showing impressive abilities towards downstream zero-shot and few-shot scenarios such as open-vocabulary question and answering. Some recent works apply LLM as a powerful knowledge base to facilitate image classification task. I2MVFormer~\cite{VLMGPT3_2} learns multi-view semantic embeddings from LLM to boost zero-shot classification. Sachit \emph{et al.}~\cite{VLMGPT3} query the LLM for descriptive features of each class, which improves the model's generalization. However, these works overly rely on the external LLM, which is not always available in deployment/inference. Furthermore, descriptions from LLM are only tested for zero-shot scenarios, leaving the potential for few-shot learning unexplored.




\section{Conclusion}
Various prompt learning methods have been developed to better transfer the pre-trained vision-language model to downstream recognition tasks. This paper proposes a novel prompt tuning method, namely Dual-ALignd Prompt Tuning (\textsc{DuAl-PT}), where prompts are aligned with powerful large language model and local image features. 
As the first work to bridge the gap between LLMs and prompt learning, \textsc{DuAl-PT} provides fresh insight into prompt engineering problem. Comprehensive experiments have been conducted on downstream tasks, and the results reveal that \textsc{DuAl-PT} is capable of further improving the transferability of the vision-language model via distilling from LLM. 
In future, with the advances of LLMs, \textsc{DuAl-PT} is expected to work better by learning from a stronger teacher. We also plan to bridge the LLMs with prompt learning in a more elegant manner.


\begin{thebibliography}{35}
\providecommand{\natexlab}[1]{#1}

\bibitem[{Alvarez-Melis and Jaakkola(2018)}]{gromov-text}
Alvarez-Melis, D.; and Jaakkola, T.~S. 2018.
\newblock Gromov-Wasserstein alignment of word embedding spaces.
\newblock \emph{arXiv preprint arXiv:1809.00013}.

\bibitem[{Bossard, Guillaumin, and Van~Gool(2014)}]{food}
Bossard, L.; Guillaumin, M.; and Van~Gool, L. 2014.
\newblock Food-101--mining discriminative components with random forests.
\newblock In \emph{Computer Vision--ECCV 2014: 13th European Conference,
  Zurich, Switzerland, September 6-12, 2014, Proceedings, Part VI 13},
  446--461. Springer.

\bibitem[{Brown et~al.(2020)Brown, Mann, Ryder, Subbiah, Kaplan, Dhariwal,
  Neelakantan, Shyam, Sastry, Askell et~al.}]{gpt3}
Brown, T.; Mann, B.; Ryder, N.; Subbiah, M.; Kaplan, J.~D.; Dhariwal, P.;
  Neelakantan, A.; Shyam, P.; Sastry, G.; Askell, A.; et~al. 2020.
\newblock Language models are few-shot learners.
\newblock \emph{Advances in neural information processing systems}, 33:
  1877--1901.

\bibitem[{Chen et~al.(2023)Chen, Yao, Song, Li, Rao, and Zhang}]{plot}
Chen, G.; Yao, W.; Song, X.; Li, X.; Rao, Y.; and Zhang, K. 2023.
\newblock {PLOT}: Prompt Learning with Optimal Transport for Vision-Language
  Models.
\newblock In \emph{The Eleventh International Conference on Learning
  Representations}.

\bibitem[{Chowdhery et~al.(2022)Chowdhery, Narang, Devlin, Bosma, Mishra,
  Roberts, Barham, Chung, Sutton, Gehrmann et~al.}]{palm}
Chowdhery, A.; Narang, S.; Devlin, J.; Bosma, M.; Mishra, G.; Roberts, A.;
  Barham, P.; Chung, H.~W.; Sutton, C.; Gehrmann, S.; et~al. 2022.
\newblock Palm: Scaling language modeling with pathways.
\newblock \emph{arXiv preprint arXiv:2204.02311}.

\bibitem[{Cimpoi et~al.(2014)Cimpoi, Maji, Kokkinos, Mohamed, and
  Vedaldi}]{dtd}
Cimpoi, M.; Maji, S.; Kokkinos, I.; Mohamed, S.; and Vedaldi, A. 2014.
\newblock Describing textures in the wild.
\newblock In \emph{Proceedings of the IEEE conference on computer vision and
  pattern recognition}, 3606--3613.

\bibitem[{Cuturi(2013)}]{sinkhorn}
Cuturi, M. 2013.
\newblock Sinkhorn distances: Lightspeed computation of optimal transport.
\newblock \emph{Advances in neural information processing systems}, 26.

\bibitem[{Deng et~al.(2009)Deng, Dong, Socher, Li, Li, and Fei-Fei}]{imagenet}
Deng, J.; Dong, W.; Socher, R.; Li, L.-J.; Li, K.; and Fei-Fei, L. 2009.
\newblock Imagenet: A large-scale hierarchical image database.
\newblock In \emph{2009 IEEE conference on computer vision and pattern
  recognition}, 248--255. Ieee.

\bibitem[{Desai and Johnson(2021)}]{virtex}
Desai, K.; and Johnson, J. 2021.
\newblock Virtex: Learning visual representations from textual annotations.
\newblock In \emph{Proceedings of the IEEE/CVF conference on computer vision
  and pattern recognition}, 11162--11173.

\bibitem[{Fei-Fei, Fergus, and Perona(2004)}]{caltech101}
Fei-Fei, L.; Fergus, R.; and Perona, P. 2004.
\newblock Learning generative visual models from few training examples: An
  incremental bayesian approach tested on 101 object categories.
\newblock In \emph{2004 conference on computer vision and pattern recognition
  workshop}, 178--178. IEEE.

\bibitem[{He et~al.(2016)He, Zhang, Ren, and Sun}]{resnet}
He, K.; Zhang, X.; Ren, S.; and Sun, J. 2016.
\newblock Deep residual learning for image recognition.
\newblock In \emph{Proceedings of the IEEE conference on computer vision and
  pattern recognition}, 770--778.

\bibitem[{Helber et~al.(2019)Helber, Bischke, Dengel, and Borth}]{eurosat}
Helber, P.; Bischke, B.; Dengel, A.; and Borth, D. 2019.
\newblock Eurosat: A novel dataset and deep learning benchmark for land use and
  land cover classification.
\newblock \emph{IEEE Journal of Selected Topics in Applied Earth Observations
  and Remote Sensing}, 12(7): 2217--2226.

\bibitem[{Jia et~al.(2021)Jia, Yang, Xia, Chen, Parekh, Pham, Le, Sung, Li, and
  Duerig}]{align}
Jia, C.; Yang, Y.; Xia, Y.; Chen, Y.-T.; Parekh, Z.; Pham, H.; Le, Q.; Sung,
  Y.-H.; Li, Z.; and Duerig, T. 2021.
\newblock Scaling up visual and vision-language representation learning with
  noisy text supervision.
\newblock In \emph{International Conference on Machine Learning}, 4904--4916.
  PMLR.

\bibitem[{Joulin et~al.(2016)Joulin, Van Der~Maaten, Jabri, and
  Vasilache}]{flickr}
Joulin, A.; Van Der~Maaten, L.; Jabri, A.; and Vasilache, N. 2016.
\newblock Learning visual features from large weakly supervised data.
\newblock In \emph{Computer Vision--ECCV 2016: 14th European Conference,
  Amsterdam, The Netherlands, October 11--14, 2016, Proceedings, Part VII 14},
  67--84. Springer.

\bibitem[{Krause et~al.(2013)Krause, Stark, Deng, and Fei-Fei}]{cars}
Krause, J.; Stark, M.; Deng, J.; and Fei-Fei, L. 2013.
\newblock 3d object representations for fine-grained categorization.
\newblock In \emph{Proceedings of the IEEE international conference on computer
  vision workshops}, 554--561.

\bibitem[{Li et~al.(2017)Li, Jabri, Joulin, and Van Der~Maaten}]{ngram}
Li, A.; Jabri, A.; Joulin, A.; and Van Der~Maaten, L. 2017.
\newblock Learning visual n-grams from web data.
\newblock In \emph{Proceedings of the IEEE International Conference on Computer
  Vision}, 4183--4192.

\bibitem[{Li et~al.(2022)Li, Li, Xiong, and Hoi}]{blip}
Li, J.; Li, D.; Xiong, C.; and Hoi, S. 2022.
\newblock Blip: Bootstrapping language-image pre-training for unified
  vision-language understanding and generation.
\newblock In \emph{International Conference on Machine Learning}, 12888--12900.
  PMLR.

\bibitem[{Li et~al.(2021)Li, Selvaraju, Gotmare, Joty, Xiong, and Hoi}]{albef}
Li, J.; Selvaraju, R.; Gotmare, A.; Joty, S.; Xiong, C.; and Hoi, S. C.~H.
  2021.
\newblock Align before fuse: Vision and language representation learning with
  momentum distillation.
\newblock \emph{Advances in neural information processing systems}, 34:
  9694--9705.

\bibitem[{Lu et~al.(2022)Lu, Liu, Zhang, Liu, and Tian}]{proda}
Lu, Y.; Liu, J.; Zhang, Y.; Liu, Y.; and Tian, X. 2022.
\newblock Prompt distribution learning.
\newblock In \emph{Proceedings of the IEEE/CVF Conference on Computer Vision
  and Pattern Recognition}, 5206--5215.

\bibitem[{Maji et~al.(2013)Maji, Rahtu, Kannala, Blaschko, and
  Vedaldi}]{aircraft}
Maji, S.; Rahtu, E.; Kannala, J.; Blaschko, M.; and Vedaldi, A. 2013.
\newblock Fine-grained visual classification of aircraft.
\newblock \emph{arXiv preprint arXiv:1306.5151}.

\bibitem[{Menon and Vondrick(2022)}]{VLMGPT3}
Menon, S.; and Vondrick, C. 2022.
\newblock Visual classification via description from large language models.
\newblock \emph{arXiv preprint arXiv:2210.07183}.

\bibitem[{Naeem et~al.(2023)Naeem, Khan, Xian, Afzal, Stricker, Van~Gool, and
  Tombari}]{VLMGPT3_2}
Naeem, M.~F.; Khan, M. G. Z.~A.; Xian, Y.; Afzal, M.~Z.; Stricker, D.;
  Van~Gool, L.; and Tombari, F. 2023.
\newblock I2MVFormer: Large Language Model Generated Multi-View Document
  Supervision for Zero-Shot Image Classification.
\newblock In \emph{Proceedings of the IEEE/CVF Conference on Computer Vision
  and Pattern Recognition}, 15169--15179.

\bibitem[{Nilsback and Zisserman(2008)}]{flowers}
Nilsback, M.-E.; and Zisserman, A. 2008.
\newblock Automated flower classification over a large number of classes.
\newblock In \emph{2008 Sixth Indian Conference on Computer Vision, Graphics \&
  Image Processing}, 722--729. IEEE.

\bibitem[{Parkhi et~al.(2012)Parkhi, Vedaldi, Zisserman, and Jawahar}]{pets}
Parkhi, O.~M.; Vedaldi, A.; Zisserman, A.; and Jawahar, C. 2012.
\newblock Cats and dogs.
\newblock In \emph{2012 IEEE conference on computer vision and pattern
  recognition}, 3498--3505. IEEE.

\bibitem[{Peyré, Cuturi, and Solomon(2016)}]{gromov-calculate}
Peyré, G.; Cuturi, M.; and Solomon, J. 2016.
\newblock Gromov-Wasserstein Averaging of Kernel and Distance Matrices.
\newblock In Balcan, M.~F.; and Weinberger, K.~Q., eds., \emph{Proceedings of
  The 33rd International Conference on Machine Learning}, volume~48 of
  \emph{Proceedings of Machine Learning Research}, 2664--2672. New York, New
  York, USA: PMLR.

\bibitem[{Pratt, Liu, and Farhadi(2022)}]{VLMGPT3_3}
Pratt, S.; Liu, R.; and Farhadi, A. 2022.
\newblock What does a platypus look like? generating customized prompts for
  zero-shot image classification.
\newblock \emph{arXiv preprint arXiv:2209.03320}.

\bibitem[{Radford et~al.(2021)Radford, Kim, Hallacy, Ramesh, Goh, Agarwal,
  Sastry, Askell, Mishkin, Clark et~al.}]{clip}
Radford, A.; Kim, J.~W.; Hallacy, C.; Ramesh, A.; Goh, G.; Agarwal, S.; Sastry,
  G.; Askell, A.; Mishkin, P.; Clark, J.; et~al. 2021.
\newblock Learning transferable visual models from natural language
  supervision.
\newblock In \emph{International conference on machine learning}, 8748--8763.
  PMLR.

\bibitem[{Radford et~al.(2018)Radford, Narasimhan, Salimans, Sutskever
  et~al.}]{gpt}
Radford, A.; Narasimhan, K.; Salimans, T.; Sutskever, I.; et~al. 2018.
\newblock Improving language understanding by generative pre-training.

\bibitem[{Radford et~al.(2019)Radford, Wu, Child, Luan, Amodei, Sutskever
  et~al.}]{radford2019}
Radford, A.; Wu, J.; Child, R.; Luan, D.; Amodei, D.; Sutskever, I.; et~al.
  2019.
\newblock Language models are unsupervised multitask learners.
\newblock \emph{OpenAI blog}, 1(8): 9.

\bibitem[{Sariyildiz, Perez, and Larlus(2020)}]{learn_caption}
Sariyildiz, M.~B.; Perez, J.; and Larlus, D. 2020.
\newblock Learning visual representations with caption annotations.
\newblock In \emph{Computer Vision--ECCV 2020: 16th European Conference,
  Glasgow, UK, August 23--28, 2020, Proceedings, Part VIII 16}, 153--170.
  Springer.

\bibitem[{Soomro, Zamir, and Shah(2012)}]{ucf101}
Soomro, K.; Zamir, A.~R.; and Shah, M. 2012.
\newblock UCF101: A dataset of 101 human actions classes from videos in the
  wild.
\newblock \emph{arXiv preprint arXiv:1212.0402}.

\bibitem[{Xiao et~al.(2010)Xiao, Hays, Ehinger, Oliva, and Torralba}]{sun}
Xiao, J.; Hays, J.; Ehinger, K.~A.; Oliva, A.; and Torralba, A. 2010.
\newblock Sun database: Large-scale scene recognition from abbey to zoo.
\newblock In \emph{2010 IEEE computer society conference on computer vision and
  pattern recognition}, 3485--3492. IEEE.

\bibitem[{Zhang et~al.(2022)Zhang, Roller, Goyal, Artetxe, Chen, Chen, Dewan,
  Diab, Li, Lin et~al.}]{opt}
Zhang, S.; Roller, S.; Goyal, N.; Artetxe, M.; Chen, M.; Chen, S.; Dewan, C.;
  Diab, M.; Li, X.; Lin, X.~V.; et~al. 2022.
\newblock Opt: Open pre-trained transformer language models.
\newblock \emph{arXiv preprint arXiv:2205.01068}.

\bibitem[{Zhou et~al.(2022{\natexlab{a}})Zhou, Yang, Loy, and Liu}]{cocoop}
Zhou, K.; Yang, J.; Loy, C.~C.; and Liu, Z. 2022{\natexlab{a}}.
\newblock Conditional prompt learning for vision-language models.
\newblock In \emph{Proceedings of the IEEE/CVF Conference on Computer Vision
  and Pattern Recognition}, 16816--16825.

\bibitem[{Zhou et~al.(2022{\natexlab{b}})Zhou, Yang, Loy, and Liu}]{coop}
Zhou, K.; Yang, J.; Loy, C.~C.; and Liu, Z. 2022{\natexlab{b}}.
\newblock Learning to prompt for vision-language models.
\newblock \emph{International Journal of Computer Vision}, 130(9): 2337--2348.

\end{thebibliography}
\end{document}